\documentclass[10pt,twocolumn,letterpaper]{article}

\usepackage{wacv}
\usepackage{times}
\usepackage{epsfig}
\usepackage{graphicx}
\usepackage{amsmath}
\usepackage{amssymb}
\usepackage{algorithm}
\usepackage{algorithmic}
\usepackage{dsfont}
\usepackage{comment}
\usepackage{multirow}
\usepackage{authblk}
\makeatletter
\renewcommand\AB@affilsepx{\quad \protect\Affilfont}
\makeatother

\def\wacvPaperID{642} 

\wacvfinalcopy 
\ifwacvfinal
\def\assignedStartPage{9876} 
\fi

\ifwacvfinal
\usepackage[breaklinks=true,bookmarks=false]{hyperref}
\else
\usepackage[pagebackref=true,breaklinks=true,colorlinks,bookmarks=false]{hyperref}
\fi

\ifwacvfinal
\else
\fi

\begin{document}

\title{SALAD: Self-Assessment Learning for Action Detection}

\author[1,3]{Guillaume Vaudaux-Ruth}
\author[1,2]{Adrien Chan-Hon-Tong}
\author[3]{Catherine Achard}
\affil[1]{ONERA}
\affil[2]{Universit\'e Paris Saclay}
\affil[3]{Sorbonne Universit\'e}
\affil[ ]{\authorcr\tt\small guillaume.vaudaux-ruth@onera.fr}
\affil[ ]{\tt\small adrien.chan\_hon\_tong@onera.fr}
\affil[ ]{\tt\small catherine.achard@sorbonne-universite.fr}




\maketitle

\begin{abstract}
   Literature on self-assessment in machine learning mainly focuses on the production of well-calibrated algorithms through consensus frameworks \textit{i.e.} calibration is seen as a problem.
   Yet, we observe that learning to be properly confident could behave like a powerful regularization and thus, could be an opportunity to improve performance.
   
   Precisely, we show that used within a framework of action detection, the learning of a self-assessment score is able to improve the whole action localization process.
   Experimental results show that our approach outperforms the state-of-the-art on two action detection benchmarks. On THUMOS14 dataset, the mAP at $tIoU@0.5$ is improved from $42.8\%$ to $44.6\%$, and from $50.4\%$ to $51.7\%$ on ActivityNet1.3  dataset. For lower tIoU values, we achieve even more significant improvements on both datasets.
\end{abstract}

\section{Introduction}
Many psychological experiments \cite{griffin1992weighing,kepecs2012computational} have shown that cross-confidence between two coworkers is strengthened by the ability of each partner to correctly evaluate the quality of his, or her, current work.
Recent work suggests that these results are also consistent with human-system cooperation \cite{berberian2010preliminary}, including systems based on machine learning.
Thus, a research direction towards even smarter machines lies in their ability to assess confidence in their own output.

Despite many works dealing with this subject, either by calibrating natural algorithm confidence \cite{ni2019calibration,ding2020revisiting} or by building a confidence model \cite{hoiem2012diagnosing,corbiere2019addressing}, it seems that this notion of correct confidence has not been intensively exploited as a way to learn the algorithm itself. 
This is even more interesting in the case of regression, where a confidence value is not easily accessible.  Thus, Gal et al. \cite{gal2016dropout} extend an old idea \cite{pomerantsev1999confidence} to estimate the confidence by collecting the results of stochastic forward passes obtained using dropout. This idea is more a pooling of results than a real self-assessment.
For the particular and difficult task of data generation, GANs allow to reach the confidence of the generated data by using a binary classification in real or synthetic data. However, such work does not allow the generator to self-assess.

\begin{figure}[t]
\includegraphics[width=0.47\textwidth]{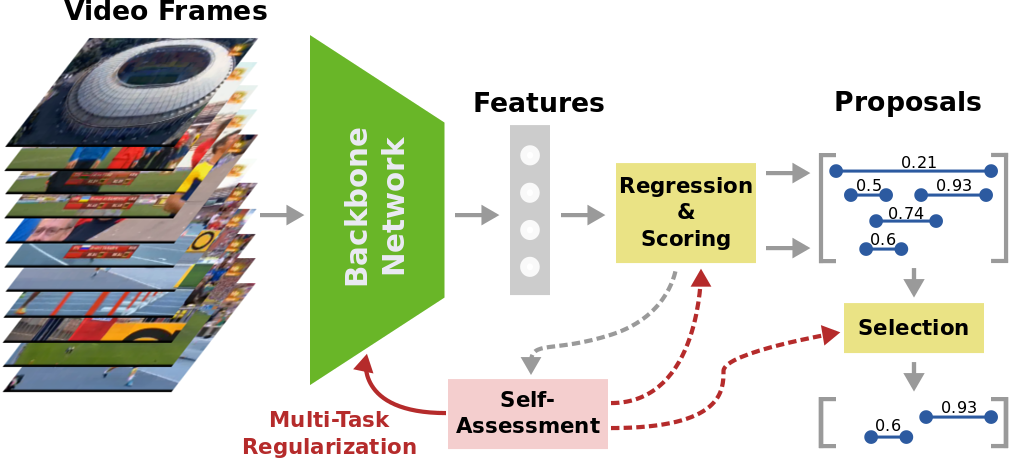}
\caption{\textbf{Joint learning of self-assessment and action localization.} At training time, self-assessment allows to improve regression, scoring and selection of temporal segments, as well as feature extraction through multi-task regularization.  \label{fig:overview1}}
\end{figure}

In this paper, we propose to simultaneously learn a regression task and a self-assessment of this task. This allows, first of all, to access a confidence in the regressed value, but above all to improve the results of the regression: for the action detection task, we outperform current state of the art \cite{xu2020gtad,DBLP:journals/corr/abs-1806-02964,DBLP:journals/corr/abs-1804-07667,DBLP:journals/corr/abs-1907-09702} by almost $2\%$ of mean Average Precision, on both THUMOS14 \cite{WangQT14} and ActivityNet1.3 \cite{caba2015activitynet} benchmarks while relying on the temporal regression of the action segments and a standard backbone. This idea is illustrated in Figure~\ref{fig:overview1}, where the self-assessment assists in feature extraction, segment regression/scoring and selection of relevant segments.

Such joint learning is particularly relevant for action detection because it allows to naturally take into account some of its specificities such as, for example, the importance of predicting only one detection per action instance (avoiding double detection). Thus, this learning optimizes both the individual confidence of each predicted segment and their mutual ability to provide a correct overall detection result. More formally, the simultaneous learning of the regression task and the self-assessment of this task allows encoding the complex mean Average Precision (mAP) metric that is not directly taken into account using a simple regression, based on local Intersection over Union (tIoU). 
Finally, this confidence can also be used to prune some frames during the learning. Such pruning, inspired by recent deep reinforcement work \cite{DBLP:journals/corr/YeungRMF15,wu2019adaframe,wu2019multi}, is naturally integrated into the proposed method as the self-assessment score could be interpreted as an attentional value.

To summarize, the main contributions of this paper are as follow: 
\begin{itemize}
    \item We show that we can simultaneously learn to be correctly confident and to regress action proposals.
    \item Such joint learning provides a confidence score on the detection and, more importantly, improves detection results by helping to find relevant features that address both tasks. 
    \item It also allows natural pruning of the frames during the learning process, resulting in improved performance. 
    \item The result is a new action detector named SALAD, which stand for \textit{Self-Assessment Learning for Action Detection}, that outperforms current state-of-the-art at $tIoU@0.5$ by $1.8\%$ on THUMOS14 and by $1.4\%$ on ActivityNet1.3. SALAD also achieve more significant improvements at lower $tIoU$.
\end{itemize}
In this paper, Section \ref{sec:related_work} presents related works. The proposed method is then introduced in Section \ref{sec:method}. The results on action detection task, and a more general discussion on the idea of confidence learning are presented in Section \ref{sec:experiments}, before the conclusion in Section \ref{sec:conclusion}.

\section{Related Work}
\label{sec:related_work}
\subsection{Video Analysis}
\textbf{Action Recognition}. Action recognition is a highly studied task in the field of video analysis. Recently, many deep learning methods have been used to improve the spatio-temporal video representations. These methods have usually used features from both RGB frames and optical flow sequences. 2D-CNNs have been used to produce those features \cite{DBLP:journals/corr/FeichtenhoferPZ16, DBLP:journals/corr/SimonyanZ14, DBLP:journals/corr/WangXW015} before the use of 3D-CNNs \cite{ DBLP:journals/corr/TranBFTP14, DBLP:journals/corr/CarreiraZ17, DBLP:journals/corr/abs-1711-10305, DBLP:journals/corr/XuDS17}. In this work we use an action recognition method to extract video features in order to use it as input to our model.

\textbf{Temporal Action Proposal.} The goal of the temporal action proposal task in untrimmed video is to detect action instances by predicting their temporal boundaries and giving them a confidence score. For proposal generation, segment-based methods \cite{DBLP:journals/corr/FeichtenhoferPZ16, gao2018ctap, scnn_shou_wang_chang_cvpr16, DBLP:journals/corr/SinghC16, Gao_2017_ICCV} or boundaries-based \cite{SSN2017ICCV, DBLP:journals/corr/abs-1806-02964, DBLP:journals/corr/abs-1907-09702, DBLP:journals/corr/abs-1811-11524, DBLP:journals/corr/YeungRMF15} methods are the most used. The first method generates direct propositions using a multi-scale anchoring at regular time intervals or a direct regression. The second method locally predicts the presence of temporal boundaries, at each temporal location in the video, and globally generates propositions by combining them. In this work, we present a segment-based method that is able to accurately regress temporal proposals. 

\textbf{Action Detection}. The objective of action detection task is to detect action instances in untrimmed videos by predicting their temporal boundaries and action categories. Two categories of action detection methods have been studied: one-stage and two-stages approaches. One-stage approaches generate both temporal proposals and their action classes \cite{sstad_buch_bmvc17, DBLP:journals/corr/YeungRMF15, DBLP:journals/corr/abs-1710-06236, huang2019decoupling}, while the two-stages approaches focus on generating temporal proposals and uses SOTA action detection method to classify the proposals \cite{DBLP:journals/corr/ShouWC16, DBLP:journals/corr/SinghC16, SSN2017ICCV, PGCN2019ICCV, DBLP:journals/corr/abs-1804-07667, DBLP:journals/corr/abs-1806-02964, DBLP:journals/corr/abs-1907-09702}. We show that we are able to generate a one-stage method capable of accurately predicting temporal proposals and their respective action class.

\subsection{Beyond just answering correctly}
The more mature machine learning becomes, the more is expected of it.
Thus, today’s machine learning modules must not only be highly accurate, but also robust to adversarial examples \cite{papernot2016limitations}, be able to deal with out-of-distribution samples \cite{vyas2018out}, be explainable \cite{bucher2018semantic}, or correctly calibrated \cite{corbiere2019addressing}.

All these functionalities, which can be seen as constraints, are now seen as opportunities because they can help design more efficient algorithms.
For example, work on adversarial defense \cite{zhang2019theoretically} have shown a trade-off between robustness and accuracy that is used to design a new defense method optimizing a regularized surrogate loss that generalizes the concept of maximum margin (of support vector machine) to deep neural network.
In another context, Yun et al. \cite{yun2020regularizing} show that encoding overall confidence is more relevant than relying on a simple local loss function that only takes into account one sample and its ground truth. Thus, they propose a regularization method that forces the network to produce more meaningful and consistent predictions and significantly improves the generalization ability and calibration performance of convolutional neural networks.

In addition to these approaches, multi-task learning \cite{zamir2018taskonomy} allows the network to jointly learn several tasks that help each other, thereby improving individual performance.

In this work, we jointly learn to produce an output and to produce a confidence score in this output. In a regression context, this allows us to exploit the opportunity to improve the regression through the regularization effect of calibration, unlike most works that focuses on confidence in a regression context \cite{pomerantsev1999confidence,gal2016dropout}.
Indeed, their work mainly considers ensemble methods to provide a measure of consensus, and not an end-to-end self-assessment learning.

\section{Proposed Method: Self-Assessement of Action Segment Regression}
\label{sec:method}
\subsection{Notations and preliminaries}
Let $\mathbf{V} = x_1, ..., x_T$ with $T$ frames (or $T$ snippets of frames) an untrimmed video  where $x_t$ denotes the feature vector of the $t$-th frame (or $t$-th snippet). This video is associated with a ground truth set of segments $\mathbf{G} = \left\{[s_{n}, e_{n}], c_n \right\}_{n=1}^{N}$, where $s_{n}$, $e_{n}$ and $c_n$ are respectively the start time, the end time and the action class of the ground-truth segment $n$.

The action detection task consist in predicting a set of $M$ proposals $\mathbf{P} = \left\{[s_{m}, e_{m}], c_m \right\}_{n=1}^{M}$ from $\mathbf{V}$ corresponding as closely as possible to $\mathbf{G}$.

\subsection{Overview of the SALAD Architecture}
\begin{figure}[t]
\includegraphics[width=0.47\textwidth]{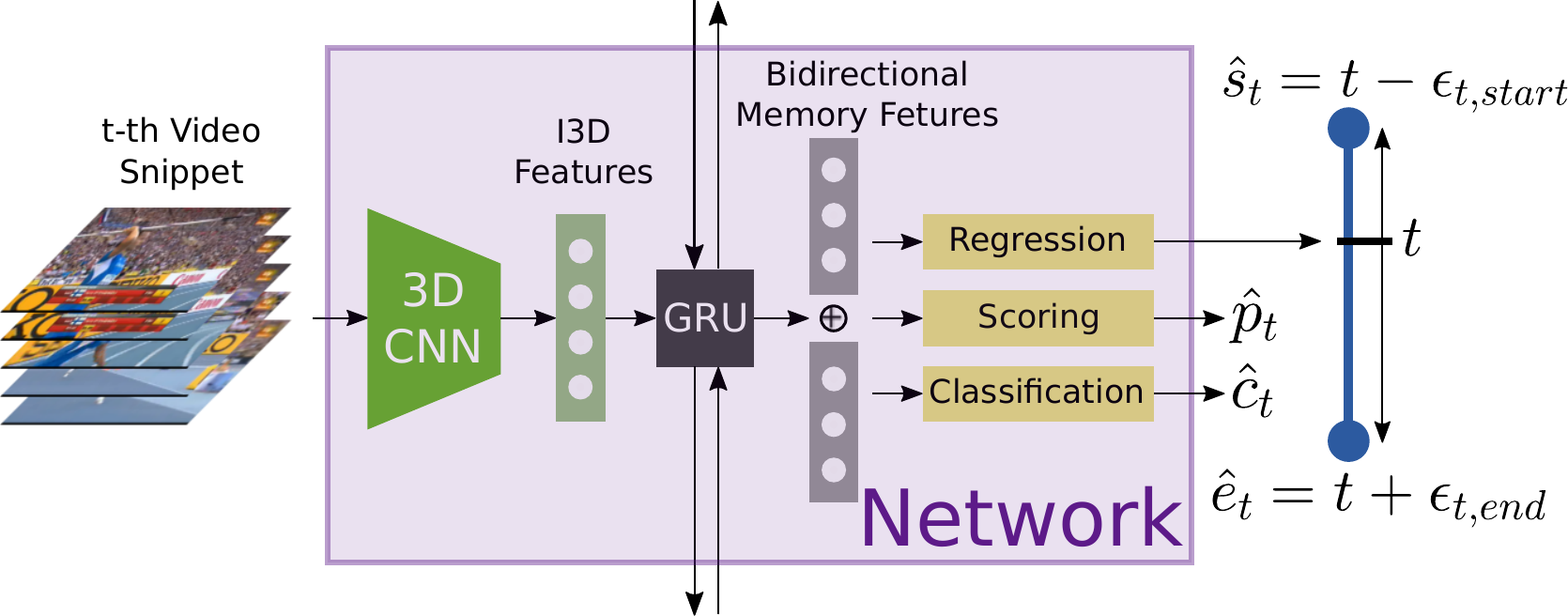}
\caption{\textbf{Architecture of SALAD network.} 
Each frame (or snippet of frames) $t$ is first represented by a feature vector using a backbone network. A bidirectional GRU is then used to produce a memory of the previous frames and a memory of the following frames. Both memories are managed by three heads that produce a proposal $[\hat{s}_t, \hat{e}_t]$ comprising the time $t$, a confidence score $\hat{p}_t$ and an action class $\hat{c}_t$.
\label{fig:network}}
\end{figure}

The overall architecture of SALAD network is presented in Figure~\ref {fig:network}. Each frame (or snippet of frames), at time $t$, is first characterized by a feature vector using a backbone network. The time series of vectors is then processed by a bidirectional Gated Recurrent Unit (GRU) which produces, for each time $t$, two latent vectors. The first one can be considered as a representation of the video before the time step $t$ while the second one is a representation of the video after this time step. The two latent vectors are then shared by 3 fully connected modules that respectively produce a regressed segment [$\hat{s}_t$, $\hat{e}_t$], a confidence score in this segment $\hat{p}_t$ and its action class $\hat{c}_t$.

But the main purpose of this article is not this architecture but the joint learning of segment regression and confidence assessment, instead of relying on an external module. The use of self-assessment allows to prune frames (or snippets) during training and the improvement of features by using an attention mechanism and a multi-task regularization, as illustrated in Figure~\ref{overall}.


Let us start by explaining the principle of joint learning in the case of a simple regression.

\subsection{ Naive regression self-assessment}
When performing a regression, only one output is available: the regressed value. It is then impossible to know how much confidence can be placed in this value. This is not the case in classification where deep network outputs distribution of scores over the classes. From this distribution, one could extract both the \textit{argmax} (the predicted class) and the margin between the max and the second max, classically used as a confidence.

For regression, one solution is to assign a score to the boxes a posteriori. Alternatively, we propose to estimate a regression confidence using a two-head network, one that performs a classical regression and the other that estimates whether or not we can be confident in this regressed value. Therefore, we express the confidence problem in the form of a binary classification.
For example, for an input $x$ associated with a ground truth $z(x)$, such a network should produce $\hat{z}(x)$ (regression) and $\hat{p}(x)$ (self-assessment). Let $\kappa$ be the allowed tolerance on the regressed value, then a loss could be:
\[l =  ||z(x)- \hat{z}(x)||_2^2 + \alpha \left[ \begin{array}{c}
     y(x) \log(\hat{p}(x)) +  \\
      (1-y(x))\log(1-\hat{p}(x))
\end{array} \right] \]
with $y(x)=1$ if  $||z(x) -  \hat{z}(x)||_2^2 < \kappa$  and $0$ otherwise\\
\textit{i.e.} a regression term and a binary cross entropy term.

Firstly, by adding such a head, the system will be able to output a confidence score which is a social requirement for real life applications.
Then, thanks to multi-task regularization, this second head could help improve regression (pre-existing idea of such framework could be found in \cite{choi2018learning}, but the benefit of self-assessment for the underlying task is not the purpose of their work).

Our contribution is to adapt such a head for action detection where the benefit is even greater because this head can be used as an attentional prior. Also, because it is easier to encode the specificity of the action detection metric in the self-assessment objective rather than in the regression itself, which can therefore lead to much better performance.

\subsection{Action detection self-assessment}
\label{sec:action_detection_self_assessment}
\begin{figure*}[t]
\centering
\includegraphics[width=0.84\textwidth]{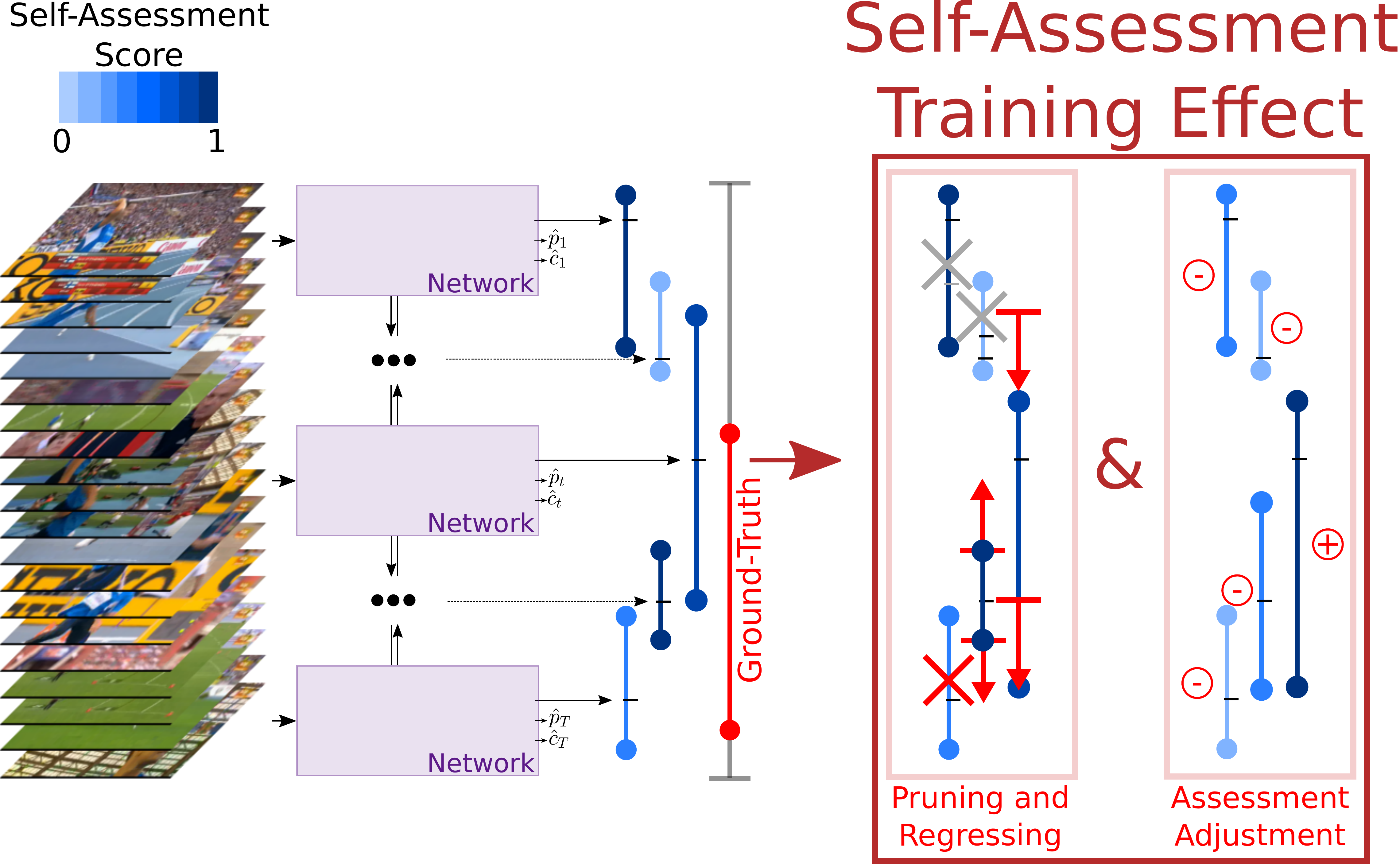}
\caption{ \textbf{Illustration of the self-assessment learning}. The regressed segments and their confidence are used, together with the ground truth, to compute the loss. During this computation, some segments are pruned (crossed out in the figure) while others continue the competition (not crossed out). The last are classified as sure or not, depending on their tIoU and their score. 
With this loss function, the regressed boundaries of the unpruned segments can evolve as close as possible to the ground truth and the confidence of best segments is optimized to increase while the others are optimized to decrease, as shown on the right of the figure with the signs \textcircled{+} and \textcircled{-}.}
\label{overall}
\end{figure*}

Action detection problem is complex as an unknown number of segments must be regressed. Moreover, we have to manage the precision of the detected segments but also the lack of detection, and double detection. 
The classical criterion used to measure the precision of the detected segments is the temporal Intersection over Union (tIoU) between the predicted segment $[\hat{s},\hat{e}]$ and the ground truth $[s, e]$, defined as
\begin{equation}
\label{eqn:iou}
tIoU( [\hat{s},\hat{e}], [s,e] ) =  \frac{\min(e,\hat{e})-\max(s,\hat{s})}{\max(e,\hat{e})-\min(s,\hat{s})}\\
\end{equation}

Classical regression heads take into account this tIoU criterion, which is estimated locally and therefore does not deal with relationships between segments, which can induce, for example, double detections.
Conversely, as in \cite{yun2020regularizing}, self-assessment could easily take into account the overall behavior. Typically, it can consider whether a regressed segment matches a ground truth with a minimum tIoU $\mu$ and whether it is the best among all other segments for that particular ground truth segment. 

Another interest of such a self-assessment head is to access the scoring during the training. Thus, this confidence could be used to prune the predicted segments on which the regression is performed, like an attentional clue. 
This pruning strategy is consistent with recent work based on reinforcement learning, which shows that the use of all frames is not optimal for detecting actions \cite{wu2019adaframe,wu2019multi}.

\textbf{Self-Assessment Loss.} At each iteration, the network predicts for each frame (or snippets) $t$, a regressed temporal segment $[\hat{s}_t, \hat{e}_t]$ containing the frame $t$ and a self-assessment score $\hat{p_t}$. 
Let us first forget about the action classes, which will be managed separately.

Then, a \textit{status} is computed for variables $(\alpha_{t, n})_{t,n\in [1,T]\times [1,N]} \in \{0,1\}$ and $(y_t)_{t\in [1,T]} \in {\{0,1\}}$: $\alpha_{t, n}$ is set to $1$ if the frame $t$ is considered to regress segment $[s_n,e_n]$, and, $y_t$ is set to $1$ if optimization should increase $p_t$.
These variables are assigned according to the Algorithm~\ref{algo1} (a more literal explanation is presented after), and are then used to compute the loss of regression and self-assessment heads using: 
\begin{equation}
\label{eq:loss}\begin{array}{c}L_{r,sa}=  \overset{T}{\underset{t=1}{\sum}}  \left(y_{t}\log(p_t)+(1-y_{t})\log(1-p_t)\right) \\ - \lambda_1 \overset{ T }{\underset{t=1}{\sum}}\overset{N}{\underset{n=1}{\sum}} \left( \alpha_{t,n} tIoU([s_n,e_n],[\hat{s}_{t},\hat{e}_{t}]) \right) \end{array} 
\end{equation}
where $\lambda_1$ is a weighting parameter.

\begin{algorithm}

\caption{Computation of the self-assessment loss\label{algo1}}
\hspace*{\algorithmicindent} 
\textbf{Input:} $\{[\hat{s}_{t}, \hat{e}_{t}], \hat{p}_{t} \}_{t=1}^T$ the segments regressed at each frame and the self-assessment value.
$\{[s_n, e_n]\}_{n=1}^N$ the ground truth action instances of the video. $\mu$ the tIoU threshold. 

\begin{algorithmic}[1]

\STATE Compute $\sigma(t)$

//series of times $t$, sorted in decreasing order of $\hat{p}_{t}$

//$p_{\sigma(t+1)}\geq p_{\sigma(t)}$

\STATE Initialize $\alpha = \mathbf{0}$, $\beta = \mathbf{0}$, $y =\mathbf{0}$
\FOR{$t=1,...,T$} 
\FOR{$n=1,...,N$}
\IF{$\beta_{n} = 0$}
\IF{$e_n\leq \sigma(t) \leq s_n$} 
\STATE $\alpha_{\sigma(t),n}\leftarrow 1$
\STATE $\rho \leftarrow tIoU([\hat{s}_{\sigma(t)}, \hat{e}_{\sigma(t)}],[s_n, e_n])$ // Eq.\ref{eqn:iou}
\IF{$\rho > \mu$}
\STATE $\beta_n \leftarrow 1$
\STATE $y_{\sigma(t)} \leftarrow 1$ 
\ENDIF
\ENDIF
\ENDIF
\ENDFOR
\ENDFOR
\STATE Compute self-assessment loss $L$ using Eq.~\ref{eq:loss} 
\end{algorithmic}
\hspace*{\algorithmicindent} \textbf{Output: } Loss $L$
\end{algorithm}

\textbf{Self-Assessment Training.}
More literally, first, we sort the frames: $\sigma$ is a permutation such that $\forall u\leq v\in [1,T], \ p_{\sigma(u)}\geq p_{\sigma(v)} $.
Then, a frame $\sigma(t)$ outside of all ground truth segments is never used for regression and is expected to lead to low self-assessment $y_{\sigma(t)}=0$, $(\alpha_{\sigma(t),n})_n=\mathbf{0}$. Inversely, let us consider a frame $\sigma(t)$ inside a ground truth segment $[s_n,e_n]$. Two cases appear. 
If this ground truth segment has already been matched with another regressed segment ($\beta_n = 1$), then it means that self-assessment $\hat{p}_{\sigma(t)}$ is lower than the one which has matched with $[s_n,e_n]$ (the loop on the frames is done by decreasing order of $\hat{p}$).
So, we do not use this frame $\sigma(t)$ for regression (it is pruned) and set $y_{\sigma(t)}=0$ in order to decrease $\hat{p}_{\sigma(t)}$. 
In the other case, the frame is considered as competitive ($(\alpha_{\sigma(t),n})_n=\mathbf{1}$) and participate to the regression loss. 
Moreover, if the tIoU between its regressed segment $([\hat{s}_{\sigma(t)}, \hat{e}_{\sigma(t)}]$ and the ground truth $[s_n, e_n])$ is higher than $\mu$, then the frame $\sigma(t)$ is considered as the better one to predict the ground truth segment ($\beta_n = 1$) and $y_{\sigma(t)}=1$ in order to try to increase $\hat{p}_{\sigma(t)}$.

It is therefore a dynamic process where, at the beginning, all frames participate in the regression of action boundaries. Then, gradually, some frames with a poor potential are pruned in order to focus the regression on pertinent predictions and to improve them. 
In addition, self-assessment is also evolving: while confidence in non-optimal frames is encouraged to decrease, confidence in the best frames (which match the ground truth with an tIoU higher than $\mu$ and have the greatest confidence in the corresponding segment) is induced to increase.

This process is illustrated Figure~\ref{overall} where five frames produce a regressed segment around it [$\hat{s_t}$,$\hat{e_t}$], a confidence $\hat{p_t}$ and a class $\hat{c_t}$. The confidence is represented by a more or less dark blue, according to the scale shown in the upper left corner of the figure. The proposals having a null $tIoU$ with the ground truth segment are pruned using grey cross, while those pruned by the self-assessment process are pruned using a red cross.

\textbf{Classification.} The classification is performed in parallel with the process. We chose to perform a frame-level classification. So at each time step, the network output $\hat{c}_t$ a probability distribution over the action classes, including a background class. We maximize the recall of the action classes over the videos using the loss:
\begin{equation}
    L_{cls} = \overset{T}{\underset{t=1}{\sum}}  w_t\left(c_t\log(\hat{c}_t)+(1-c_t)\log(1-\hat{c}_t)\right)
\end{equation}
where $w_t$ is set to $0$ if $c_t$ is background, and $1$ otherwise.
\textbf{SALAD Loss function.} The overall loss used for training is the sum of the regression/self-assessment loss and the classification loss (with a $\lambda_2$ weighting parameter):
\begin{equation}
    L = L_{r,sa} + \lambda_2 L_{cls} 
\end{equation}

Importantly, there is a classification loss at training, but we do not use the classification confidence (see ablation study in Section \ref{sect:classification}). Indeed, we find that classification is good, but poorly calibrated.
So there is no point in merging classification confidence (essentially random) with our regression self-assessment score(which is crucial here for action detection).

\section{Experiments}
In this section, we first discuss the datasets and the details of our implementation. Then, SALAD is compared to state-of-the-art approaches. Finally, we examine the contribution of each component of our self-assessment learning to the task of action detection through ablation studies and discussions.
\label{sec:experiments}

\subsection{Datasets}
We evaluate our approach on two challenging datasets:

\textbf{THUMOS14} \cite{WangQT14} that contains 410 untrimmed videos with temporal annotations for 20 action classes. Training and validation sets include respectively 200 and 210 videos and each video has more than 15 action instances.

\textbf{ActivityNet1.3} \cite{caba2015activitynet} that contains 19,994 videos with 200 action classes collected from YouTube. The dataset is divided into three subsets: 10,024 training videos, 4,926 validation videos and 5,044 testing videos.

\subsection{Implementation details}

\begin{table}[]
    \centering
    \begin{tabular}{c c c c c}
    \hline
    \multirow{2}{*}{Module} &  \multirow{2}{*}{Layer} & Input &  Output &  \multirow{2}{*}{Activation}  \\
     & & Size & Size & \\
    \hline
     \hline
    Memory & GRU & 2048 & 2048 & Identity \\
     \hline
    \multirow{4}{*}{Regression} & Linear & 2048 & 2048 & ReLu \\
     & Linear & 2048 & 1024 & ReLu \\
     & Linear & 1024 & 1024 & ReLu \\
     & Linear & 1024 & 2 & Sigmoid \\
     \hline
     \multirow{4}{*}{Scoring} & Linear & 2048 & 2048 & ReLu \\
     & Linear & 2048 & 1024 & ReLu \\
     & Linear & 1024 & 1024 & ReLu \\
     & Linear & 1024 & 1 & Sigmoid \\
     \hline
     \multirow{3}{*}{Classification} & Linear & 2048 & 2048 & ReLu \\
     & Linear & 2048 & 1024 & ReLu \\
     & Linear & 1024 & classes + 1 & Softmax \\
     \hline
    \end{tabular}
    \caption{The detailed architecture of the network.}
    \label{tab:network}
\end{table}

\textbf{Detection Metric}. The common practice in action detection is to use the mean Average Precision (mAP) at different tIoU thresholds to evaluate the quality of a set of detections. Following previous work, the tIoU thresholds $\{0.1, 0.2, 0.3, 0.4, 0.5\}$ and $\{0.5, 0.75, 0.95\}$ are respectively used for THUMOS14 and ActivityNet1.3.

\textbf{Features}. For both datasets, we use the same backbone network to extract features. As in the most recent work \cite{DBLP:journals/corr/abs-1804-07667, PGCN2019ICCV}, we use the two-stream features, extracted by I3D network \cite{DBLP:journals/corr/CarreiraZ17}, pre-trained on Kinetics. We use the pre-extracted features provided by \cite{paul2018w}. The videos are previously sampled at 25 frames per second, and TV-L1 optical flow algorithm \cite{TVL1} is applied. From that, the features are extracted from non-overlapping 16-frame video slices to produce 2 feature vectors of size 1024 (RGB and Flow).

\textbf{Network Construction}. The network architecture, presented Figure~\ref{fig:network}, has been designed with the parameters in Table~\ref{tab:network}. RGB and Flow stream are computed together, the features are then fused in a 2048 feature vector. Thus, we keep this size for each GRU latent vector. The three network heads, for regression, scoring and classification, have respectively 4, 4 and 3 fully connected layers whose number of neurons are given in  Table~\ref{tab:network}. The boundary regression is done relatively to the position $t$, as presented in Fig.\ref{fig:network}. Thus, for a simplified implementation, the regression head produces a normalized version of $[\epsilon_{t, start}, \epsilon_{t, end}]$.

\textbf{Training and Inference.} We implement our framework using Pytorch 1.0, Python 3.7 and CUDA 10.0. The optimization is done using Adam, with an initial learning rate of $10^{-4}$. For THUMOS14 dataset, we set the batch size to 4 and for ActivityNet1.3, we set it to 16. Both datasets are trained during 100 epochs. Note that the convergence is better when a pre-training of the classification head is done before the whole training, done using $\lambda_1 = 1$, $\lambda_2 = 0.1$ and $\mu = 0.5$. During inference, we use all the proposals produced by the network, and soft-NMS for computing mAPs on THUMOS14 (one per time step) while we use a maximum of 20 proposals for ActivityNet1.3 since the number of ground-truth per video is lower.

\subsection{Comparison with state-of-the-art results}
\textbf{THUMOS14.} Table~\ref{tab:thumos14} compares our model with state-of-the-art detectors on THUMOS14 dataset. The proposed method achieves the highest mAP for all thresholds, implying that the self-assessment process is capable in producing very accurate proposals. Especially, our method outperforms the previous best performance reported at tIoU@0.1 by more than 7\% and improves the mAP at tIoU@0.5 from 42.2\% to 44.6\%. We also combine our method with P-GCN\cite{PGCN2019ICCV}, the current state-of-the-art post-processing method. This combination slighly improves our results at every tIoU and outperforms all state of the art methods at tIoU $<0.5$. These results also show  that our self-assessment learning does not require as much post-processing as other methods whose scores deteriorate without it.

\begin{table}[]
    \centering
    \begin{tabular}{l c c c c c}
         \hline
         tIoU & 0.1 & 0.2 & 0.3 & 0.4 & 0.5\\
         \hline
         Oneata \textit{et al.} \cite{oneata:hal-01074442}& 36.6 & 33.6 & 27.0 & 20.8 & 14.4 \\
         Wang \textit{et al.} \cite{WangQT14} & 18.2 & 17.0 & 14.0 & 11.7 & 8.3 \\
         Caba \textit{et al.} \cite{Heilbron16}& - & - & - & - & 13.5 \\
         Richard \textit{et al.} \cite{7780706}& 39.7 & 35.7 & 30.0 & 23.2 & 15.2 \\
         Shou \textit{et al.} \cite{DBLP:journals/corr/ShouWC16}& 47.7 & 43.5 & 36.3 & 28.7 & 19.0 \\
         Yeung \textit{et al.} \cite{DBLP:journals/corr/YeungRMF15}& 48.9 & 44.0 & 36.0 & 26.4 & 17.1 \\
         Yuan \textit{et al.} \cite{7780710}& 51.4 & 42.6 & 33.6 & 26.1 & 18.8 \\
         DAPs \cite{DAPs}& - & - & - & - & 13.9 \\
         SST \cite{sst_buch_cvpr17}& - & - & 37.8 & - & 23.0 \\
         CDC \cite{cdc_shou_cvpr17}& - & - & 40.1 & 29.4 & 23.3 \\
         Yuan \textit{et al.} \cite{yuan17}& 51.0 & 45.2 & 36.5 & 27.8 & 17.8 \\
         SS-TAD \cite{sstad_buch_bmvc17}& - & - & 45.7 & - & 29.2 \\
         CBR \cite{gao2017cascaded}& 60.1 & 56.7 & 50.1 & 41.3 & 31.0 \\
         Hou \textit{et al.} \cite{Hou2017}& 51.3 & - & 43.7 & - & 22.0 \\
         TCN \cite{Dai_2017_ICCV}& - & - & - & 33.3 & 25.6 \\
         TURN-TAP \cite{Gao_2017_ICCV}& 54.0 & 50.9 & 44.1 & 34.9 & 25.6 \\
         R-C3D \cite{DBLP:journals/corr/XuDS17}& 54.5 & 51.5 & 44.8 & 35.6 & 28.9 \\
         SSN \cite{SSN2017ICCV} & 66.0 & 59.4 & 51.9 & 41.0 & 29.8 \\
         BSN \cite{DBLP:journals/corr/abs-1806-02964}& - & - & 53.5 & 45.0 & 36.9 \\
         BMN \cite{DBLP:journals/corr/abs-1907-09702}& - & - & 56.0 & 47.4 & 38.8 \\
         Chao \textit{et al.} \cite{DBLP:journals/corr/abs-1804-07667}& 59.8 & 57.1 & 53.2 & 48.5 & 42.8 \\
         G-TAD \cite{xu2020gtad} & - & - & 54.5 & 47.6 & 40.2 \\\hline
         \textbf{SALAD} & \textbf{73.3} & \textbf{70.7} & \textbf{65.7} & \textbf{57.0} & \textbf{44.6} \\\hline
         BSN + PGCN \cite{PGCN2019ICCV} & 69.5 & 67.8 & 63.6 & 57.8 & 49.1 \\
         G-TAD + PGCN & - & - & 66.4 & 60.4 & \textbf{51.6} \\\hline
         \textbf{SALAD} + PGCN  & \textbf{75.2} & \textbf{73.4} & \textbf{69.4} & \textbf{61.6} & 49.8 \\\hline
         \\
    \end{tabular}
    \caption{\textbf{Action detection results on testing set of THUMOS14}, measured by mAP (\%) at different tIoU thresholds. SALAD significantly outperforms all the other methods for all IoU and is even slightly improved by a P-GCN combination.}
    \label{tab:thumos14}
\end{table}

\begin{table}[]
    \centering
    \begin{tabular}{l c c c c}
    \hline
         tIoU & 0.5 & 0.75 & 0.95 & Average  \\\hline
         Singh \textit{et al.} \cite{DBLP:journals/corr/SinghC16} & 34.47 & - & - & - \\
         SCC \cite{8099821} & 40.00 & 17.90 & 4.70 & 21.70 \\ 
         CDC \cite{cdc_shou_cvpr17} & 45.30 & 26.00 & 0.20 & 23.80 \\ 
         R-C3D \cite{DBLP:journals/corr/XuDS17} & 26.80 & - & - & - \\ 
         SSN \cite{SSN2017ICCV} & 39.12 & 23.48 & 5.49 & 23.98 \\  
         BSN \cite{DBLP:journals/corr/abs-1806-02964} & 46.45 & 29.96 & 8.02 & 30.03 \\ 
         Chao \textit{et al.} \cite{DBLP:journals/corr/abs-1804-07667} & 38.23 & 18.30 & 1.30 & 20.22 \\ 
         P-GCN \cite{PGCN2019ICCV} & 48.26 & 33.16 & 3.27 & 31.11 \\ 
         BMN \cite{DBLP:journals/corr/abs-1907-09702} & 50.07 & \textbf{34.78} & 8.29 & 33.85 \\ 
         G-TAD \cite{xu2020gtad} & 50.36 & 34.60 & \textbf{9.02} & \textbf{34.09} \\\hline
         \textbf{SALAD} & \textbf{51.72} & 31.21 & 3.33 & 31.02 \\\hline
    \end{tabular}
    \caption{\textbf{Action detection results on validation set of ActivityNet1.3}, measured by mAP (\%) at different tIoU thresholds and the average mAP. SALAD achieves the best performance for IoU@0.5.} 
    \label{tab:activitynet13}
\end{table}

\begin{table}[]
    \centering
    \begin{tabular}{l c c c c c}
         \hline
         tIoU & 0.1 & 0.2 & 0.3 & 0.4 & 0.5\\\hline
         BMN \cite{DBLP:journals/corr/abs-1907-09702} & 70.91 & 64.46 & 58.79 & 54.14 & 50.07 \\
         \textbf{SALAD} & \textbf{77.68} & \textbf{70.66} & \textbf{64.06} & \textbf{57.45} & \textbf{51.72}\\\hline
    \end{tabular}
    \caption{\textbf{Action detection results on validation set of ActivityNet1.3}, measured by mAP (\%) for lower tIoU thresholds than 0.5. SALAD significantly outperforms BMN at low tIoU.}
    \label{tab:joker}
\end{table}
\textbf{ActivityNet1.3.} Tab.\ref{tab:activitynet13} reports the state-of-the-art results on ActivityNet1.3 dataset. Our algorithm outperforms the previous best performance at tIoU@0.5 by 1.4\%. At higher tIoU, some state-of-the-art methods are more efficient than ours. 



However, any state-of-the-art algorithm performs poorly at high tIoU values, especially for 0.95 where the best algorithm only reaches 9\% of mAP. Thus, we consider that all current methods are not sufficiently mature to handle action detection at high tIoU.

It is not surprising that self-assessment is not useful on such ambiguous problem: the more ambiguous a problem is, the less it is possible to have a consistent self-assessment.

Now, even on ActivityNet, we are improving the mAP at tIoU $<0.5$. Although these results are not conventionally reported, we compare SALAD at lower IoU with BMN \cite{DBLP:journals/corr/abs-1907-09702}, which is the best open-sourced method available at the time of writing (\textit{JJBOY/BMN-Boundary-Matching-Network}). SALAD results, presented in Table~\ref{tab:joker}, clearly outperform those of BMN, for all tIoU $\leq 0.5$.

As soon as the detection problem is well posed and unambiguous, SALAD algorithm outperforms the state-of-the-art methods on both THUMOS14 and ActivityNet1.3 datasets.

\subsection{Ablation Studies / Discussions}
The original purpose of this article is to jointly learn segment regression and segment scoring through self-assessment. 
Such learning leads, in particular, to an improvement in the features quality and thus to an increase in performance as shown before.

In this section, we decompose the proposed loss in order to understand which parts are important for the improvement of the results and to better understand what leads to this significant improvement. We continue to not use frames outside of the ground truth segments (as in Section \ref{sec:action_detection_self_assessment}).

\textbf{Pruning.} We first investigate the influence of pruning. Regarding Equation~\ref{eq:loss}, it consists in quantifying the influence of the $\alpha_{t,n}$ parameters.
Precisely, we compare five methods. (\textit{i}) No pruning ($(\alpha_{t,n})_{t,n}=\mathbf{1}$). (\textit{ii}) Random pruning where $(\alpha_{t,n})$ values are randomly set to 0 or 1.
(\textit{iii}) Top 1 IoU pruning where only the proposal with the best IoU is regressed.
(\textit{iv}) Frozen pruning that consists in extracting the final $(\alpha_{t,n})$ values from SALAD and relearning the algorithm from the beginning with these frozen values.
(\textit{v}) SALAD.

Results, presented in Table~\ref{tab:pruning}, clearly show that pruning is an import key to the algorithm, since too much pruning (top 1) leads to very bad results and no pruning or random pruning to bad results.
This is consistent with previous work as \cite{wu2019adaframe,wu2019multi} which also shows that pruning is important.

More importantly, by using the frozen pruning from our best model, performance is worse. 
So, regression alone allows to obtain similar results because it seems important to adapt pruning to the current behavior of the algorithm.
Thus, even if pruning is a key component, it is not the only one, which highlighting the relevance of self-assessment. 


\textbf{Learning self-assessment score.} We then investigate the influence of self-assessment ($y_t$ parameter in Equation~2) and compare different ideas.
The first is to ignore the IoU  and set $y_t=1$ for the frame with the highest $\hat{p}_t$ inside each ground truth segment.
The second idea is that any segment with a confidence level $\hat{p}_t$ greater than $0.5$ is considered sure. The last idea imposes a condition on the IoU, that should be greater than the threshold $\mu$ (without considering $\hat{p}_t$).
Thus, the last two methods do not take into account neighboring segments when assigning the self-assessment. 

\begin{table}[]
    \centering
    \begin{tabular}{l c c c c c}
    \hline
    mAP@tIoU & 0.1 & 0.2 & 0.3 & 0.4 & 0.5\\\hline
    No Pruning & 66.2 & 63.0 & 57.0 & 46.9 & 32.0\\
    Top 1 IoU & 55.7 & 53.4 & 48.3 & 38.9 & 27.4 \\
    Random & 65.8 & 62.6 & 56.6 & 46.2 & 32.9  \\
    Frozen & 63.2 & 57.5 & 45.4 & 45.4 & 31.5  \\
    SALAD (pruning) & \textbf{73.3} & \textbf{70.7} & \textbf{65.7} & \textbf{57.0} & \textbf{44.6}\\
    \hline
    \end{tabular}
    \caption{Comparison between our SALAD training and different pruning and regression strategies on THUMOS14, measured by mAP(\%)}
    \label{tab:pruning}
\end{table}

\begin{table}[]
    \centering
    \begin{tabular}{l c c c c c}
    \hline
    mAP@tIoU & 0.1 & 0.2 & 0.3 & 0.4 & 0.5\\\hline
    $y_t=1 \Leftrightarrow t=\sigma(0)$ & 59.4 & 56.7 & 51.3 & 42.0 & 30.6 \\
    $y_t = 1 \Leftrightarrow \hat{p}_t > 0.5$ & 66.4 & 62.7 & 53.8 & 41.2 & 26.7 \\
    $y_t = 1 \Leftrightarrow tIoU_t > \mu$ & 65.5 & 62.3 & 53.7 & 40.9 & 28.0  \\
    SALAD & \textbf{73.3} & \textbf{70.7} & \textbf{65.7} & \textbf{57.0} & \textbf{44.6}\\
    \hline
    \end{tabular}
    \caption{Comparison between SALAD and other self-assessment strategies on THUMOS14, measured by mAP(\%).}
    \label{tab:self-assessment}
\end{table}

\begin{figure*}[t]
\centering
\includegraphics[width=0.9\textwidth]{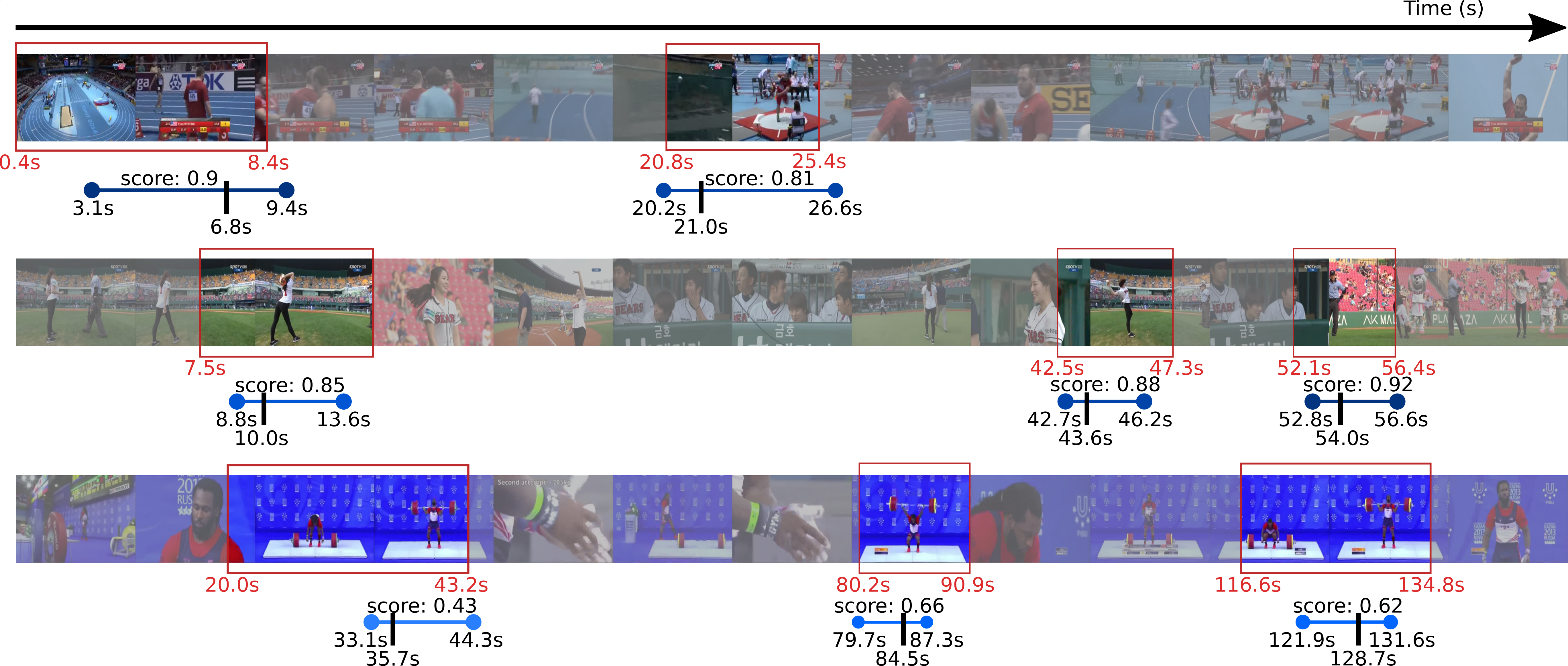}
\caption{\textbf{Qualitative results.} We show qualitative localization results on THUMOS14 dataset. Ground-truth segments are red boxes. The predictions made by SALAD are blue segments and the time of prediction are the black time steps.}
\label{results}
\end{figure*}
The results, presented in Table~\ref{tab:self-assessment}, show that the three alternative assignments of $y_t$ lead to a dramatic drop in performance compared to SALAD.
They highlight that the relevance of self-assessment comes from the introduction of action detection specificities during the learning process, such as, for example, allowing just one predicted segment per ground truth.
So, for a given ground truth segment, the regression, for example, of the segment with the best confidence, degrades the results. Similarly, the regression of segments with an high confidence, without taking into account other segments, or the regression of only the segments with good IoU, are not optimal.


The previous ablation studies on pruning and self-assessment learning clearly establish that SALAD success comes from its ability to prune frames during training and from the self-assessment process that allows to inject prior on the mAP metric that can hardly be injected with classical local loss. 

\subsection{Classification confidence}
\label{sect:classification}
In all the experiments presented before, the natural classification confidence is discarded (classification head outputs a probability distribution over the classes from which a naive level of confidence can be obtained).
However, it is questionable whether it would be relevant to merge the two confidences. Let us note $p_r$ and $p_c$ the regression score and classification confidence.

Thus, we show in Table~\ref{tab:fusion} a comparison with different fusion strategies: the arithmetic mean of the two confidences $\frac{p_r+p_c}{2}$, their geometric product $\sqrt{p_r p_c}$ and the product of the regression confidence with a normalized classification confidence $p_r\times (1-\exp(-\zeta p_c))$. The idea of this last method is to decrease regression confidence in case of important ambiguity in classification. Finally, SALAD confidence is only the regression confidence: $p_r+0\times p_c$.

\begin{table}[]
    \centering
    \begin{tabular}{l c c c c c}
    \hline
    mAP@tIoU & 0.1 & 0.2 & 0.3 & 0.4 & 0.5\\\hline
    Arithmetic mean & 63.5 & 61.7 & 58.2 & 51.1 & 41.1 \\
    Geometric mean & 65.7 & 63.8 & 60.0 & 52.7 & 42.4  \\
    Normalized product  & 67.2 & 65.2 & 64.5 & 53.8 & 43.2  \\
    SALAD & \textbf{73.3} & \textbf{70.7} & \textbf{65.7} & \textbf{57.0} & \textbf{44.6}\\
    \hline
    \end{tabular}
    \caption{Comparison between SALAD and confidence fusion methods on THUMOS14, measured by mAP(\%).}
    \label{tab:fusion}
\end{table}

Results clearly show that it is not relevant to merge our self-assessment confidence with that of classification, highlighting why SALAD does not.

\subsection{Qualitative Results}
In Figure~\ref{results}, we present some localization results on THUMOS14 dataset. Of course, the display of some samples provides limited information about the overall behavior of our algorithm.
However, a very important point we want to highlight is that the frame used to regress segment is uniformly distributed across the ground truth segment.
It is therefore very important to select the best frames during the learning, as proposed in SALAD.

\section{Conclusion}
\label{sec:conclusion}
In this paper, we propose a new action detection algorithm named SALAD that outperforms states-of-the-art on both THUMOS14 and ActivityNet1.3 datasets.

This performance gain is achieved by adding self-assessment directly into the network learning.
Indeed, this self-assessment allows to prune frames (or snippets of frames) and to improve features by using attentional mechanism and multi-task regularization.

In addition, this self-assessment allows to capture all the specificity of action detection metric in the loss function, contrary to regression losses which only measure local performance. Thus, contrary to the common opinion that robustness, calibration or explainability are considered constraints, we introduce one of them during the learning process as a way to improve performance.
\\

\textbf{Acknowledgments:} This work was performed using HPC resources from GENCI-IDRIS (2019-AD011011269)

{\small
\bibliographystyle{ieee_fullname}
\bibliography{egbib}
}

\end{document}